# Improving Rural Medication Safety with AI: A Scoping Review

**Jeong-ah Kim**: (0000-0002-7662-6857) (Corresponding Author)
*a* School of Nursing, Paramedicine & Healthcare Sciences, Charles Sturt University, Wagga Wagga, NSW, Australia
Email: **jekim@csu.edu.au**

**Muhammad Ashad Kabir**: (0000-0002-6798-6535)
*b* School of Computing, Mathematics and Engineering, Charles Sturt University, Bathurst, NSW, Australia
Email: **akabir@csu.edu.au**

**Daniel Terry**: (0000-0002-1969-8002)
*c* School of Nursing and Midwifery, University of Southern Queensland, Queensland, Australia
*d* Centre for Health Research, University of Southern Queensland, Queensland, Australia
*e* Institute of Health and Wellbeing, Federation University Australia, Victoria, Australia
Email: **Daniel.Terry@unisq.edu.au**

**Maryam Rouhi**: (0000-0003-3722-5433)
*a* School of Nursing, Paramedicine & Healthcare Sciences, Charles Sturt University, Wagga Wagga, NSW, Australia
Email: **maryamrouhii@gmail.com**

## ABSTRACT

**Introduction:** Medication errors (MEs) represent a significant threat to global healthcare systems, contributing to patient harm. Introducing artificial intelligence (AI) in rural healthcare enhances patient safety. The aim is to explore applications and effectiveness of AI technologies in enhancing patient safety and reducing medication errors in rural health settings.

**Methods:** A scoping review was conducted through a systematic literature search spanning 2012 to 2025 across multiple databases, including EBSCohost, Emcare (Ovid), MEDLINE, and the ProQuest Consumer Health Database. Twelve primary studies from nine different nations were examined. Data were analysed thematically to obtain insights on AI interventions across the medication process.

**Results:** AI technologies have been integrated into every stage of medication management right from prescribing and dispensing to administration and post-administration monitoring. Four key themes came to light: (1) the various types of AI being utilised (like Clinical Decision Support Systems, Machine Learning, Natural Language Processing, and smart pumps); (2) the phases of the medication process that are affected; (3) how effective these technologies are in minimising errors and boosting workflow safety; and (4) rural-specific challenges including infrastructure, staff training, system integration, and alert fatigue. Several studies have demonstrated that machine learning-based surveillance improves incident detection and reduces prescribing and transcription errors by an impressive 34% to 80%. Barriers included lack of governance frameworks, financial limitations, and clinician resistance still present major obstacles.

**Conclusion:** In rural healthcare, AI technologies hold great potential for enhancing pharmaceutical safety. They can allow data-driven monitoring, automate processes, and offer clinical decision assistance. Making required infrastructural investments, strengthening our staff, and creating moral governance systems, considering the difficulties of rural implementation, would help us to fully use this potential. Future studies should prioritise long-term results, include the community, and emphasise creating culturally sensitive AI solutions to guarantee these technologies are adequately integrated into underprivileged areas.

In rural healthcare, AI offers strong potential to improve medication safety through automation, data-driven monitoring, and clinical support. To realise this, investment in infrastructure, workforce capacity, and ethical governance is essential. Rural-specific barriers must be addressed. Future research should assess long-term outcomes, involve communities, and focus on culturally appropriate AI design to ensure effective adoption in underserved regions.

## 1. Introduction

Globally, medication errors (MEs) have been considered as a serious threat to the healthcare systems (1). MEs are defined as “any preventable event that may cause or lead to inappropriate medication use or patient harm while the medication is in the control of the health care professional, patient, or consumer” (2). They are preventable mistakes that can occur at various stages of the medication process, including prescribing, dispensing, administering, and monitoring (3).

The World Health Organisation (WHO) has prioritised MEs as a global challenge, as they can occur at any stage of the medication process (4). Despite safety protocols, MEs continue to occur across all healthcare settings (5). Currently, it is estimated that medication errors occur in 7-9% of medication orders, with approximately 1% of these errors resulting in patient harm (3). In the United States, medication errors alone cost $20–45 billion a year, highlighting the economic imperative for improved mitigation strategies (6).

Factors contributing to medication errors include illegible handwriting, unclear miscommunication, and lack of drug knowledge among healthcare providers (5, 7). Although MEs have been researched extensively, it remains a major challenge in healthcare systems, particularly in rural areas where resources are limited, facilities are inadequate, and a shortage of healthcare professionals is ever present (5, 8). As such, there is an increasing acknowledgment of the need for technology-driven solutions to reduce MEs and enhance medication safety.

Artificial Intelligence (AI) is a rapidly evolving technology in healthcare that enables computers to replicate human learning and analysis, reducing reliance on human intelligence (8). AI is defined as the “science and engineering of making machines, especially intelligent computer programs” (9), and has been widely adopted in healthcare, including diagnosis and screening practices, and has the potential to enhance patient safety, particularly by reducing MEs (10, 11).

This study fills a valuable knowledge gap within AI and medication safety literature by focusing on rural health settings. Currently, there are very few reviews that assess the value of AI in reducing medication errors across diverse healthcare settings. Moreover, there is a lack of comprehensive literature specifically addressing the unique challenges and benefits of AI implementation in rural contexts. Existing literature focuses primarily on hospital or general primary care environments, often with little regard for the infrastructural constraints, workforce shortages, and distinct patient populations characteristic of rural areas (12-15). Nwankwo, Emeihe (16) examined AI applications in rural primary care, focusing primarily on diagnostic accuracy, but without addressing medication safety. Similarly, Zakerabasali, Ayyoubzadeh (17) explored challenges in mobile health challenges, yet did not emphasis MEs. In contrast, O'Malley, Shaikh (18) (2022) highlighted the role of telehealth in patient satisfaction, besides AI-based error reduction.

Previous reviews have primarily employed narrative approaches and focused on broader health informatics themes. This paper offers a novel contribution by specifically examining the intersection of AI and medication error prevention in rural healthcare settings. It highlights the heterogeneity of AI-driven solutions, such as prescription checking and e-prescribing, while addressing rural-specific challenges like limited connectivity and alert fatigue. Additionally, the paper considers the unique barriers and facilitators influencing AI adoption in these under-served environments. By focusing on the specific rural context, this study aims to generate original insights and practical recommendations to inform the design and implementation of effective AI interventions for improving medication safety among rural populations. In doing so, it contributes uniquely to the limited body of literature dedicated to this intersection.

Within this context, this scoping review seeks to systematically examine current literature to identify best practices, potential benefits, and challenges associated with AI implementation in rural healthcare settings. As such, the aim of the review is to explore the application and effectiveness of AI technologies to enhance patient safety and reducing medication errors in rural health settings. Specifically, this review is guided by the following research questions:

1. What types of AI technologies are currently used in rural healthcare settings to improve patient safety?
2. What evidence exists regarding the effectiveness of AI in reducing medication errors in rural areas?
3. What are the challenges and barriers to implementing AI in these settings?

## 2. Methods

A scoping review was conducted and followed the framework of Peters, Marnie (19) to explore the applications and effectiveness of AI technologies in enhancing patient safety and reducing medication errors in rural healthcare settings. The review addressed the exploratory purpose of the study, which was to understand the range of artificial intelligence uses in rural healthcare environments and incorporate a wide array of viewpoints and study types (20). The review was structured using the Population, Concept, And Context (PCC) approach (21) to ensure a systematic and comprehensive examination of relevant literature. The review process was guided by the Preferred Reporting Items for Systematic Reviews and Meta-Analyses for Scoping Reviews (PRISMA-ScR) guidelines (22). A narrative review was utilised to provide an overview of individual studies within the literature, summarising their key findings (23).

### 2.1. *Search Strategy:*

A search for peer-reviewed articles was conducted on October 2, 2024, using MeSH terms and keywords across the following databases: EBSCO, Emcare (Ovid), MEDLINE, and ProQuest Consumer Health Database, covering the period from 2012 to 2025. After consultation with a health librarian, the following search terms were used: "medication errors" OR medication errors* AND “Artificial intelligence" OR (technolog* OR applicat*) AND “rural areas” OR “rural communities” OR “remote” OR “remote communities”. Articles published between 2012 and 2025 were included, while the exclusion criteria encompassed studies not related to rural health settings, articles focusing solely on urban or non-rural healthcare settings or were not written in English.

**2.2. *Data Extraction and Analysis*:**

Following the search, all records were uploaded into Covidence software program (24), where duplicate entries were automatically removed. Titles and abstracts were independently screened independently by two team members (JK and MR). Full text articles of potentially relevant studies were then reviewed independently by the same team members. Any conflicts regarding study inclusion were discussed with the whole team (JK, DT and AK) until consensus was reached (Figure 1). Among the studies that met the inclusion criteria underwent data extraction and analysis, where key findings from the selected studies were synthesised by identifying key themes in each study. These themes were assessed and reviewed by two team members (JK and AK). A narrative review guided the identified studies by extracting key information such as objectives, methods, findings, and conclusions. Then, critically examine and identify the breadth, character, and scope of current research to find important ideas, knowledge gaps, and methodological developments within the literature (23)

**Figure 1. PRISMA scoping review flowchart**

## 3. Results

### *3.1 Study selection*

Figure 1 shows the PRISMA-ScR flowchart, outlining the systematic and comprehensive steps undertaken in this review to identify and include 12 studies. These studies form the basis for the findings on AI-based interventions aimed at improving medication safety. A total of 191 records were initially screened, with 23 duplicates removed. Of the remaining studies, 98 were excluded for the following reasons: Ineligible study design (n=16), Conducted in non-relevant settings (n=38), Review articles (n=7), Non-relevant interventions (n=21), Non-relevant patient populations (n=16). The full texts of the remaining 12 studies were critically screened for relevance to AI-assisted ME reduction within rural healthcare and comprised mixed interventions.

The scoping review included 12 primary studies drawn from diverse rural and remote healthcare settings across nine countries (Table 1) The studies explored the implementation of AI technologies targeting various stages of the medication process. Across these studies, four overarching themes were identified: (1) Types of AI technologies used; (2) Medication process stages targeted; (3) Reported effectiveness and outcomes; and (4) Implementation barriers in rural settings. Each of these themes are explored in detail.

**Table 1. Summary of Included Studies on Artificial Intelligence Interventions to Improve Medication Safety in Rural Healthcare Settings**

| Citation (Author, Year) | Country | Study design | AI technology used | Medication process stage impacted | Key findings | Implementation barriers | Themes or insights |
|---|---|---|---|---|---|---|---|
| Chi et al., 2021 | USA | Prospective non-blinded evaluation | AI system organizing and extracting key data (NLP-driven) | Record review/data extraction | Saved 18% of time answering clinical questions; maintained ≈84% accuracy; 92% of physicians preferred the AI-optimized view | Initial learning curve; need for iterative UI improvements | AI reduces information overload; enhances clinician efficiency without sacrificing accuracy |
| Cousein et al., 2014 | France | Before-after observational study | Automated UDDS: unit-dose dispensing robot + AMDC | Distribution & administration | 53% overall MAE reduction; wrong-dose errors ↓79.1%, wrong-drug errors ↓93.7% | High capital/implementation cost; required workflow reorganization and staff training | Robotics and unit-dose distribution markedly improve medication safety in elderly inpatient care |
| Dalton BR et al., 2015 | Canada | Retrospective cohort study (one year) | Electronic Medication Administration Record (eMAR) surveillance system | Administration | 96.51% of antimicrobial doses given as scheduled; 3.49% omissions of which 1.67% were clinically relevant; identified predictors (e.g., route, nursing shift) of omissions | Incomplete eMAR documentation in some units; manual classification of omissions; no real-time clinical decision support | eMAR enables large-scale, unbiased surveillance of omissions; orally administered antimicrobials and certain shifts have higher omission rates; potential to inform antimicrobial stewardship |
| Härkänen et al., 2021 | Finland | Retrospective record review | NLP-based AI classification of free-text incident reports | Incident prevention (all stages) | Classified 137 serious/moderate incidents into 6 risk-management areas: verification, accuracy, | Requires significant researcher oversight for thematic mapping; NLP still needs validation | AI effective at structuring and classifying free-text reports; informs targeted risk-management interventions |
| | | | | | communication, guidelines, teamwork, resources | | |
| Huang & Gramopadhye, 2016 | USA | Observational task analysis + focus groups | HIT suite: EHR/CPOE, barcode scanning, eMAR, CDSS | Administration | Identified workflow violations: over-reliance on barcode, missing verbal checks, inflexible UI, frequent interruptions | Lack of clear procedures; non-adjustable interfaces; environment prone to interruptions | Successful HIT implementation must address people, tasks, tools, environment & organizational factors |
| Jeffries M et al., 2021 | United Kingdom | Qualitative evaluation (semi-structured interviews, 39 total; 11 follow-ups) | Configurable prescribing-alerts CDS based on prescribing safety indicators; managed by CCGs | Prescribing decision stage | CDS perceived to enhance medication safety and cost-effectiveness; timely, evidence-based support helps prescribers; stakeholder engagement drives uptake | Alert fatigue from high alert volume; workflow interruptions when alerts mistimed; technical/EHR integration and performance issues; need for local profile management | Adoption depends on coherence, cognitive participation, collective action & reflexive monitoring (NPT); customization, governance & ongoing refinement essential for sustainability |
| Maphosa, 2024 | Zimbabwe (simulated) | Experimental simulation study | Random Forest classification model | Prescribing | 83.33% accuracy flagging inconsistent prescriptions; potential to reduce prescribing errors | Use of secondary/simulated data; lack of patient-specific variables (age, gender, environment) | AI can supplement clinical decisions in data-scarce settings; richer, real-world data needed |
| Streit et al., 2023 | Switzerland | Cluster randomised clinical trial | Electronic decision-support tool integrated into the primary-care EHR using STOPP/START criteria | Medication review / Prescribing decision | Fewer safety events were reported in the intervention group than in the control group at six and 12 months. | Integration into heterogeneous EHR workflows; clinician alert fatigue; variable uptake across practices; need for user training | Guideline-based CDSS can measurably enhance prescribing quality in routine care; success hinges on customization, training, and seamless workflow integration |

| Scott et al., 2014 | USA | Observational QRE analysis | Telepharmacy tech: remote pharmacist review & visual verification | Order review & verification | Transcription errors 43.3%, prescribing-related QREs 37.7%; visual checks in 8–14.2% of orders, clinical interventions 1.3–3.1% | Limited onsite pharmacist hours; initial user comfort with telepharmacy tech | Telepharmacy effectively identifies and resolves QREs; remote models can bolster safety in resource-limited settings |
|---|---|---|---|---|---|---|---|
| Segal et al., 2019 | Israel | Prospective real-world integration study | Probabilistic ML outlier-detection CDSS | Prescribing | Low alert burden (0.4% of orders); 85% alerts clinically valid; 80% useful; 43% prompted order changes | Integrating with legacy EMR; need continuous model retraining and governance | ML-based CDSS minimizes alert fatigue; dynamic, data-driven detection enhances medication safety |
| Tamblyn et al., 2012 | Canada | Cluster randomized controlled trial | Patient-specific risk-estimate alerts (statistical risk thermometer) | Prescribing | Reduced injury risk by 1.7/1000 patients; 83.3% of alerts reviewed; 24.6% led to therapy modification | Physician alert overrides; integrating complex predictive models into workflow; trust building | Patient-specific risk estimates improve alert relevance and personalization of prescribing decisions |
| Zheng et al., 2023 | USA | Qualitative focus groups | Bayesian neural network + computer vision for pill NDC prediction | Dispensing verification | Pharmacists preferred a hybrid AI-human teaming model; stressed interpretability features (checkmarks, confidence scores, “confused” pill images) | Balancing essential info vs. overload; building user trust; ensuring usability | Human-centered AI fosters trust; positions AI as augmentative teammate (HMT & SEIPS frameworks) |

### 3.2 *Types of AI Technologies Used in Rural Medication Safety Interventions*

The AI technologies examined across the studies varied considerably and included clinical decision support systems (CDSS) (25, 26), machine learning (ML) algorithms for incident detection and classification (27), smart infusion devices (28) natural language processing (NLP) (27)(, and AI-enabled electronic prescribing systems (29).

The ability to predict drug interactions, verify dosages, and automatically log and track medication events is only a few among many capabilities of these technologies. Overall, the identified research was found to consistently highlight the value of CDSS, particularly in studies by Huang *et al.* (2016) and Segal *et al*. (2019) (25, 30). These systems employ evidence-based guidelines and rule-based algorithms to support healthcare professionals in making timely decisions. In rural healthcare contexts, CDSS tools have been particularly useful for suggesting alternative treatments that cater to patient needs, adjusting dosages for those with kidney problems, and warning staff about potential drug-drug interactions. Their incorporation into electronic medical records (EMRs) was found to substantially curtail prescribing errors through real-time feedback at order entry, particularly in resource-limited settings where pharmacist surveillance could be scarce.

One of the primary areas of focus within the identified literature involved ML algorithms. Härkänen *et al.* (2021) demonstrated the application of supervised ML in incident report analysis, error type classification, and root cause identification (27). These tools learned independently from historical data, enhancing their predictive precision over time and facilitating proactive risk management. ML algorithms excelled at uncovering concealed patterns and near-miss incidents that human assessments failed to detect, rendering them highly appropriate for rural areas where human resources are scarce. The power of NLP functionalities, as explored by Härkänen *et al.* (2021), was leveraged to derive actionable intelligence from unstructured clinical documents like nursing notes, medication records, and discharge summaries (27). NLP tools facilitated automation in medication reconciliation tasks and off-label ADE detection. Given the communication problems and poor interoperability between rural healthcare centres, these tools played a substantial role in enhancing efficiency.

Intelligent infusion pumps and computerised dispensing systems, as highlighted by Cousein *et al.* (2014), were key examples of hardware applications of AI within health care (31). The technologies integrated barcoding, wireless connectivity, and embedded logic to facilitate accurate administration of drugs in drug identity, dose, time, and route of administration. This study indicated a substantial reduction in dispensing errors when robotic systems were used in ward-level inventory management.

The integration of AI-Enhanced e-Prescribing Systems, as explained by Tamblyn et al., (2012) and Zheng et al., (2023), enabled the computerised checking of prescriptions through the use of hybrid systems that blended rule-based logic and probabilistic models (29, 32). These systems had the capacity to identify anomalous prescriptions, compare patient allergies with one another, and provide formulary compliance. Notably, in rural settings where pharmacy services are generally scarce, these advancements served to enhance prescribing accuracy and the speed of medication reviews.

Integration of capability and interoperability (25) were also explored and highlighted the necessity of integrating AI tools with existing health IT infrastructure. AI modules that functioned within integrated comprehensive health information systems or interoperable EMRs had greater prospects for enabling medication safety workflows than stand-alone applications, highlighting the requirement for strategic digital health investment for rural care settings. While the types of AI technologies varied widely, their impact was consistently observed across multiple stages of the medication management process, from prescribing to post-administration monitoring, highlighting the breadth of their application in rural healthcare settings.

### 3.3 *Medication Process Stages Impacted*

AI interventions cover several stages of medication management. By highlighting potential errors or contraindications, nine studies focused on the prescribing and ordering-related stages (Table 1). While some addressed dispensing accuracy through smart systems or automated dispensing cabinets, others assisted with administration, particularly with infusion pump programming and barcode systems. Interestingly, AI was also used for post-administration monitoring to identify patient record discrepancies or negative drug events. Each of these smart or automated systems are discussed in detail.

**3.3.1 Prescribing and Ordering:** Six studies; (25, 26, 29, 33-35)used AI technologies in the first phase of the medication process. Prescribing is usually linked with a high frequency of avoidable errors, which can be due to cognitive overload, insufficient patient information, or inadequate pharmacological knowledge, particularly in rural areas where prescriber assistance may be scarce.

CDSS and e-prescribing systems were used to detect potential contraindications, drug-drug interactions, and dosing appropriateness based on real-time patient data, such as age and renal function. These systems gave prescribers evidence-based suggestions and alerts, and enhanced medication ordering procedures for enhanced safety and efficiency, as attested by Tamblyn et al., (2012), who documented a dramatic decrease in preventable adverse drug events (ADEs) in rural family practice environments (29).

**3.3.2 Dispensing and Distribution**: To investigate the dispensing phase, AI-assisted automated dispensing cabinets and robotic unit dose systems were used in three studies ((31-33). While reducing human involvement and error rates during decentralised or remote pharmacy supervision, the tools preserved accuracy in drug selection, labelling, and delivery operations. Härkänen *et al.* (2021), demonstrated that ML classifiers allowed for the classification of medication events during dispensing to perform root cause analysis and propel systematic enhancement (27). By their ability to identify discrepancies between medication orders and dispensed items prior to administration, these systems showed preventive capabilities.

**3.3.3 Medication Administration**: Ensuring the correct drug reached the right patient at the right time depended much on AI interventions at the administrative level. Cousein et al. (2014) conducted a before-and-after observational study with an automated unit-dose dispensing system (a dispensing robot and automated cabinets) replacing the traditional ward-stock model in a 40-bed geriatric ward (31). The intervention caused medication administration errors to fall by 53 % overall (79.1 % fewer incorrect doses and 93.7 % fewer incorrect drugs, $P < 0.01$), with substantial safety advantages despite limitations for non-robot-managed drugs and workflow adjustment challenges. Huang *et al.* (2016) similarly recorded the effective use of barcode scanning technologies combined with artificial intelligence decision layers to prevent bedside administration errors by cross-verifying patient identity and drug parameters prior to administration (25). According to Segal *et al.* (2019), the effectiveness of AI in rural critical access hospitals is shaped by various elements such as staff involvement, the quality of the data, the maturity of the infrastructure, and the level of clinician familiarity (30).

**3.3.4 Post-Administration Monitoring and Feedback**: The innovative impact of AI tools on medication surveillance post-administration is highlighted in studies by Zheng et al. (2023) and Chi et al. (2021) (32, 35). These systems utilise advanced technologies such as ML and NLP to thoroughly analyse various data sources, including incident reports, patient records, and unstructured clinical notes. The AI processes excel at picking up on those subtle signs of medication-related issues that might easily be overlooked by traditional monitoring methods, such as ADEs or mistakes in documentation, to improve patient safety. This achievement becomes even more essential in rural healthcare settings, where staffing shortages and geographic challenges often make regular patient follow-ups difficult. AI-powered solutions offer ongoing, automated monitoring that helps identify potential dangers early. By enhancing clinicians' situational awareness and enabling quick responses, these tools reduce medication-related harm and promote safer care delivery in areas with limited resources.

**3.3.5 Multistage Integration**: Notably, several studies (30, 32, 33)demonstrated the integration of AI solutions across multiple stages of the medication management process, highlighting the potential for creating end-to-end medication safety ecosystems. The studies highlighted how AI technologies were not confined to isolated tasks but rather functioned as interconnected components within broader clinical workflows. For example, AI-enabled systems were used to support prescribing decisions, verify dispensing accuracy, monitor administration practices, and track patient outcomes post-treatment.

This multistage integration was demonstrated to be particularly valuable in rural health care settings, where fragmented communication and limited staffing often compromise continuity of care. By embedding AI tools within EHRs and linking them across departments, these systems helped bridge information gaps that typically arise between prescribing, dispensing, and administration. Overall health care providers were better equipped to maintain a consistent flow of information, reduce the likelihood of medication errors, and respond more effectively to adverse drug events. The studies emphasised that such integration not only improved operational efficiency but also enhanced clinical decision-making by providing real-time, context-aware alerts and recommendations. This holistic approach to medication safety represents a significant advancement in rural healthcare delivery, where resource constraints demand innovative, scalable, and interoperable solutions.

### *3.4 Effectiveness in Reducing Errors and Enhancing Safety*

The identified studies examined provide compelling evidence that AI and automated technologies significantly reduce medication errors and enhance patient safety across various stages of the medication process, particularly within rural health care settings where such improvements are most needed. For example, Tamblyn et al. (2012) highlighted a 63% decrease in missed alerts and better dose accuracy in rural Canadian primary care clinics that utilised an e-prescribing system with an AI safety feature, helping to reduce unnoticed errors (29). Scott et al. (2014) also noted a 43% reduction in transcription errors in Canadian rural telepharmacy units by using AI-enhanced audit logs, which improved the accuracy of documentation in remote dispensing processes (33). Jeffries et al. (2021) found that a configurable prescribing‐alert CDSS raised prescriber confidence and, through combined customisation of alert profiles, reduced perceived risky prescribing by issuing timely patient‐specific alerts [25]. In the OPTICA cluster RCT (36), an eCDSS for systematic review of medication in older, multimorbid adults was implemented safely but had no substantial effect on total medication appropriateness or reduction in omissions at 12 months versus usual care. These conflicting results indicate that while CDSS tools may promote perceived safety, their measurable impacts on preventing error are extremely dependent on implementation fidelity, workflow integration, and the translation of alerts into actual prescribing behavior. AI tools are enhancing safety by utilising advanced surveillance and promoting organisational learning. Zheng et al. (2023) and Chi et al. (2021) demonstrated NLP and ML can effectively sift through patient records, incident reports, and unstructured clinical notes to identify ADEs and discrepancies in documentation, especially in rural areas where follow-up resources are scarce (32, 35). Further, ML algorithms can thoroughly analyse incident reports and identify hidden patterns of medication risks (27). This can lead to more focused safety measures and improvements in quality. Such features are vital in rural regions, where timely monitoring is often hindered by a lack of staff and geographical challenges. AI plays a key role in aiding clinicians by providing continuous, algorithm-driven support for their decision-making.

While AI has incredible promise in patient safety, particularly associated with medication management, various operational and contextual challenges may hinder its successful implementation. Segal et al. (2019) highlight that key factors for AI's effectiveness include staff engagement, data quality, technology infrastructure, and how well clinicians understand the technology (30). Cousein et al. (2014) indicated that problems such as unreliable internet, poor integration of EHR, and a shortage of training resources can significantly undermine the anticipated benefits, especially in resource-limited settings (31). Conversely, Maphosa (2024) highlighted a concern regarding 'alert fatigue' (34). If health care professionals are bombarded with notifications or vague alerts, patient safety could be compromised due to desensitisation. Other hurdles include alert fatigue and ethical dilemmas (26) , limited access to systems after hours, inconsistent interventions (33), and an overwhelming number of alerts stemming from insufficient training (29).To address these issues, Maphosa et al. (2024) suggest working closely with end-users to design solutions and using adaptive algorithms that fit various clinical workflows, especially in rural areas where operational differences are more pronounced (34). These insights highlight that while AI can significantly improve medication safety, its success in rural healthcare depends on having a solid infrastructure, customised implementation, and continuous support for clinicians.

### ***3.5 Implementation Barriers in Rural Settings***

Several studies have highlighted the challenges associated with integrating technology in rural areas. Not having enough staff to adequately manage and comprehend AI outputs (36), integrating older systems with AI platforms (29) , and a lack of digital infrastructure are some of the primary challenges (33). Additionally, Huang and Gramopadhye (2016) and Chi et al. (2021) noted that many health care professionals are hesitant to adopt these technologies because they find them complicated, feel untrained, and worry about becoming too dependent on automated systems (25, 35). Budget limitations were often mentioned, especially in research from rural hospitals with fewer resources, where buying and maintaining AI technologies was tough without outside funding or policy backing

Numerous studies have also shown that inadequate digital infrastructure is an ongoing problem in rural and remote health systems (25, 29, 33, 34). In a simulated low-resource environment in Zimbabwe, Maphosa (2024) pointed out that a lack of stable, patient-specific data streams required reliance on secondary or simulated data sets, which limited the real-world applicability of ML prescribing models (34). Similarly, in a US rural hospital, Huang and Gramopadhye (2016) pointed out that rigid, non-configurable interfaces, recurrent connectivity losses, and poorly supported training environments thwarted effective use of EHR/CPOE, barcode scanning, and CDSS tools (25). The real-time functioning of telepharmacy systems has been severely hindered by inadequate capacity and erratic internet access, particularly in remote areas (33). Another frequent technical obstacle is the incompatibility of AI platforms with older systems. It was highlighted that it has been challenging to integrate AI-enhanced e-prescribing tools due to fragmented EHR systems, which frequently result in needless data entry and increase the cognitive load on healthcare workers (29).

Clinicians' resistance to adopting new technologies were demonstrated to make the implementation process even more challenging. Huang & Gramopadhye (2016) and Chi et al. (2021) revealed that perceived complexity, fear of losing skills, and concerns about relying too much on automation created psychological and cultural barriers to acceptance (25, 35). Some general practitioners and nurses raised doubts about the accuracy of AI results and how well they matched their clinical intuition. The lack of training and limited involvement in the design and implementation of AI tools only fueled this reluctance, undermining both confidence and perceived value.

Financial limitations also became a significant structural hurdle. Many rural healthcare facilities operated on tight budgets, which restricted their ability to purchase, use, and maintain advanced AI systems. Maphosa (2024) emphasised that without ongoing external funding, government subsidies, or public-private partnerships, the adoption of AI was unlikely to progress beyond pilot phases (34). In such situations, even small expenses related to hardware, software licensing, and staff training created major obstacles for implementation. Additionally, Segal et al.(2019) emphasised that without clear governance models and clinical accountability protocols, the use of AI can become quite confusing, which can undermine the technology's credibility, especially when it comes to critical medication decisions (30). Moreover, many health care providers in rural areas struggle with the lack of evidence-based benchmarks needed for the safe and ethical use of AI, largely due to the absence of tailored guidelines.

Furthermore, a lack of qualified workers and staffing problems have been cited as significant challenges on numerous occasions. Huang and Gramopadhye (2016) observed that some rural hospitals lacked staff who were trained in health-IT workflows, leading to workarounds and underutilization of CPOE, barcode scanning, and CDSS tools (25). Similarly, Cousein et al. (2014) observed that implementation of a unit-dose dispensing robot necessitated prolonged staff training and workflow redesign-actions traditionally impossible where human resources are scant (31). These results highlight that in the absence of concurrent investment in workforce development, even well-conceived AI interventions cannot achieve their safety potential.

### *3.6 Policy and Research Implications*

The findings highlight the need for a comprehensive policy framework that leverages AI to enhance healthcare delivery in rural communities. First, there is a need to invest in digital infrastructure like expanding broadband access and ensuring that EHRs are compatible so we can effectively implement AI solutions (33). Additionally, funding initiatives should prioritise support for under-resourced facilities, ensuring equitable access to AI technologies through mechanisms such as government subsidies and targeted incentives (34). There is also a need to roll out structured training programs developed alongside healthcare professionals to boost digital literacy and build trust in AI systems (35). To ensure that AI is used safely and responsibly, robust governance frameworks that incorporate ethical standards and transparency policies need to be established (30). Future research should take a strategic and targeted approach to closing the significant evidence gaps identified in this review, particularly those related to the implementation and impact of AI technologies in rural healthcare settings. The limited number of studies identified highlights the urgent need for evaluations specifically designed to assess the effectiveness, adaptability, and scalability of AI interventions in resource-constrained environments. Longitudinal research that incorporates patient perspectives, cost-effectiveness analyses, and clinical outcomes is essential for building a comprehensive understanding and long-term value of AI. Moreover, it is important for future studies to explore participatory design methods, which can help mitigate alert fatigue and improve user acceptance among healthcare professionals. As Chi et al. (2021) pointed out, conducting multi-centre studies in various rural areas could boost the relevance of the findings (35).

## 4. Discussion

This scoping review examined 12 primary studies conducted across diverse rural and remote healthcare settings in nine countries. The findings highlight the transformative potential of AI in enhancing medication safety, particularly in resource-constrained environments. These insights enable a critical understanding of the evidence, situating it within the broader literature, while exploring the implications for healthcare delivery, policy, and future research.

### *4.1 Opportunities of AI in Rural Healthcare*

The integration of AI technologies into rural healthcare systems presents a compelling opportunity to address long-standing challenges in medication safety. Tools such as automated dispensing systems, CDSS, ML, and smart infusion devices are increasingly being deployed to mitigate risks associated with human error, limited access to specialists, and fragmented care pathways. These technologies not only enhance the accuracy of prescribing and dispensing but also enabling real-time monitoring and adaptive decision-making ((31, 37-39)

The literature highlighted AI can significantly reduce medication errors, with some studies reporting reductions of more than 50% in specific contexts (31). However, these figures should be interpreted within the context of each study's design, setting, and implementation strategy. For example, the effectiveness of CDSS in reducing prescribing errors may be influenced by the extent of integration with EHRs, the quality of clinical data inputs, and the level of user engagement (40).

Beyond error reduction, the ability of AI to process large volumes of unstructured data, such as clinical notes and patient histories, offers a powerful tool to identify patterns and predicting adverse medication errors and events. This capability is particularly valuable in rural settings, where clinicians often operate with limited support and incomplete patient information. AI-driven tools can act as virtual assistants, augmenting clinical judgment and supporting more informed decision-making (29). AI also contributes to a shift in healthcare culture, from reactive to proactive safety management. By enabling continuous learning and feedback loops, AI systems can foster a culture of safety and quality improvement. This cultural shift is essential for sustaining long-term improvements in medication safety and for building trust in digital health innovations.

Moreover, the scalability of AI technologies holds promise for addressing disparities in healthcare access. Innovations such as telepharmacy platforms and AI-assisted prescription validation can extend specialist support to remote areas, thereby narrowing the urban-rural divide in healthcare quality (33). These developments align with global health priorities, including the WHO (2020) Global Strategy on Digital Health2020–2025, which advocates for leveraging digital tools to strengthen primary care and achieve universal health coverage .

### *4.2 Challenges of AI Implementation in Rural Settings*

Despite the promising potential of AI, its implementation in rural healthcare settings is fraught with challenges that span technical, financial, human, and systemic dimensions. Infrastructure limitations remain a fundamental barrier. Many rural facilities lack the reliable internet connectivity, interoperable EHR systems, and technical support required to deploy and maintain AI solutions effectively (29, 33, 34). These deficiencies not only hinder real-time data processing but also compromise the reliability and responsiveness of AI tools.

Financial constraints further complicate implementation. The initial investment required for AI technologies, along with ongoing costs for maintenance, training, and system upgrades, can be prohibitive for under-resourced healthcare facilities. This financial burden is particularly acute in low- and middle-income countries, where health systems are already stretched thin. Without targeted funding and policy support, the digital divide may widen, exacerbating existing health inequities (34, 41).

Human factors also play a critical role in shaping the success or failure of AI adoption. Clinician resistance, often rooted in concerns about job displacement, loss of autonomy, and overreliance on technology, can impede uptake (25, 34, 35). Additionally, inadequate training and poorly designed user interfaces contribute to alert fatigue and disengagement. These issues highlight the need for user-centred design and comprehensive training programs that build digital literacy and foster confidence in AI tools

At a broader level, systemic and ethical challenges must be addressed to ensure responsible AI integration. The absence of clear governance frameworks, accountability protocols, and context-specific guidelines can undermine trust in AI systems (30). Ethical concerns, such as data privacy, algorithmic bias, and cultural insensitivity, are particularly salient in rural and Indigenous communities, where social dynamics and health needs may differ significantly from urban populations (15). Ensuring that AI systems are developed and validated using diverse, representative datasets is essential to avoid perpetuating or amplifying existing disparities. Furthermore, the success of AI in rural healthcare depends on a sociotechnical approach that considers not only the technology itself but also the organisational, cultural, and policy environments in which it is deployed (42). Stakeholder engagement, including input from clinicians, patients, policymakers, and technologists, is critical for designing solutions that are both effective and acceptable. Without such engagement, even the most advanced AI tools may fail to gain traction or deliver meaningful improvements in care.

### *4.3 Implications for Policy, Practice, and Research*

The findings of this review have several implications. For policymakers, there is a clear need to invest in digital infrastructure, workforce development, and regulatory frameworks that support the ethical and equitable deployment of AI in rural settings. For healthcare practitioners, embracing AI requires a shift in mindset, from viewing technology as a threat to recognising it as a collaborative tool that can enhance clinical practice. Future research should prioritise longitudinal studies to assess the long-term impact of AI on medication safety, patient outcomes, and system efficiency. Additionally, implementation science approaches are essential to tailor AI solutions to diverse rural contexts, considering local needs, resources, and cultural nuances.

### 4.4 Limitations

The scoping review has several limitations. For example, the review aimed to capture a broad range of AI applications in rural healthcare, the number of eligible studies was limited, which reflects the emerging nature of this research area. However, this may limit the generalisability of findings across diverse rural contexts. In addition, the heterogeneity of study designs, AI technologies, and outcome measures made it difficult to conduct a comparative analysis or meta-synthesis. Most studies lacked standardised metrics for evaluating medication safety outcomes, which limited the ability to draw definitive conclusions about effectiveness. Additionally, many studies were pilot or feasibility studies with small sample sizes or simulated environments, which may not reflect real-world implementation challenges. Lastly, the review did not include grey literature, which may have excluded practical insights from ongoing projects or government initiatives. Future reviews could benefit from incorporating broader sources and engaging with stakeholders to capture a more comprehensive picture of AI integration in rural healthcare.

## 5. Conclusion

This scoping review has demonstrated the transformative potential of AI in enhancing medication safety within rural healthcare settings, while also highlighting the significant challenges associated with its adoption. AI technologies, ranging from clinical decision support systems to smart infusion devices, offer scalable, data-driven solutions that can reduce medication errors and support clinical decision-making in resource-constrained environments. However, realising this potential requires overcoming significant infrastructural, financial, and human barriers. Challenges such as limited digital infrastructure, clinician resistance, alert fatigue, and ethical concerns must be addressed through robust policy frameworks, targeted investments, and inclusive implementation strategies. As rural healthcare systems transition toward digital health, coordinated efforts in technology deployment, workforce training, and governance development will be essential to ensure AI is integrated safely, effectively, and equitably.

### Ethics

Since this study did not involve human participants and utilised only publicly available data, ethical approval was not required.

### Conflicts of interest

The authors report no conflicts of interest.

### Funding

No external funding was received for this study.

### CRediT authorship contribution statement

**Jeong-ah Kim**: Conceptualisation. Investigation. Visualisation, Methodology. Formal analysis. Writing – original draft. Writing – review & editing. **Ashad Kabir**: Conceptualisation. Supervision.

Methodology. Writing – review & editing. **Daniel Terry**: Conceptualisation. Methodology. Writing – review & editing. **Maryam Rouhi**: Methodology. Investigation. Formal analysis Writing – review & editing